\documentclass{article}
\usepackage{spconf,amsmath,graphicx,multirow,booktabs,xcolor,cite}

\title{A Region-Based Descriptor Network For Uniformly Sampled KeyPoints}
\name{Kai Lv, Zongqing Lu, Qingmin Liao\thanks{Thanks to the National Natural Science Foundation of China (61771276) for funding.}}
\address{Department of Electronic Engineering, Graduate School at Shenzhen, Tsinghua University}
\begin{document}
%
\maketitle
\begin{abstract}
Matching keypoint pairs of different images is a basic task of computer vision. Most methods require customized extremum point schemes to obtain the coordinates of feature points with high confidence, which often need complex algorithmic design or a network with higher training difficulty and also ignore the possibility that flat regions can be used as candidate regions of matching points. In this paper, we design a region-based descriptor by combining the context features of a deep network. The new descriptor can give a robust representation of a point even in flat regions. By the new descriptor, we can obtain more high confidence matching points without extremum operation. The experimental results show that our proposed method achieves a performance comparable to state-of-the-art.
\end{abstract}
\section{Introduction}
\label{sec:intro}

Matching keypoint pairs between multiple images is an important problem to Structure-from-Motion (SfM)~\cite{1}, Simultaneous Localization and Mapping (SLAM), camera pose estimation, image retrieval~\cite{2}, 3D reconstruction~\cite{3}. The general process to get sparse matching keypoint pairs is to first detect the keypoints and then calculate their descriptors. Through some distance metric of the descriptors such as the Euclidean distance and some feature matching algorithm, such as the nearest neighbor search (NNS) algorithm~\cite{4}, the keypoints can be matched.

\begin{figure}[htb]
\begin{minipage}[b]{1.0\linewidth}
  \centering
  \centerline{\includegraphics[width=8.5cm]{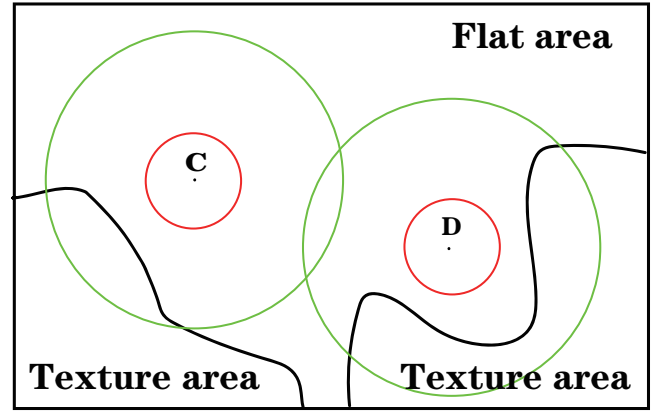}}
\end{minipage}

\caption{According to common sense, there is no key point in flat regions, like point C shown in Fig.1. The point C with a small neighbor (red circle) can not give a valid description. With the increase of the receptive field, it is possible to find a valid description for point C with a larger neighbor (green circle).}
\label{fig:res}
\end{figure}

For the hand-crafted method, the local feature extraction process is generally detect-then-describe. Firstly, detect the coordinates of the keypoints uses some custom extremum algorithm. And then according to some prior knowledge, a high-dimensional vector describing the features around the keypoint is designed. These keypoints are then matched by using these high-dimensional vectors. The best known traditional algorithm is SIFT~\cite{5}, which obtains the position of the keypoints based on the extremum detection of the complicated Difference of Gaussian. It continues to be improved. RootSIFT~\cite{11} enhances the matching accuracy through normalized descriptors, \cite{12} uses binary descriptors to improve real-time performance. \cite{13} proposes a quantization expression to provide sparse representation for image blocks. However, such algorithms usually only consider low-level information, such as corners \cite{8} or blob-like structures~\cite{5,10}. Their results deteriorate under some challenging conditions, e.g., the large variation of light intensity, sharp change of viewpoint, or weak texture area~\cite{7}. And in these challenging scenarios, they are easy to produce a lot of mismatched results, which increases the difficulty of further processing. Moreover, they often need to design a complex algorithm to complete the process of feature point detection and description.

\begin{figure*}[htb]
\begin{minipage}[b]{1.0\linewidth}
  \centering
  \centerline{\includegraphics[width=18cm]{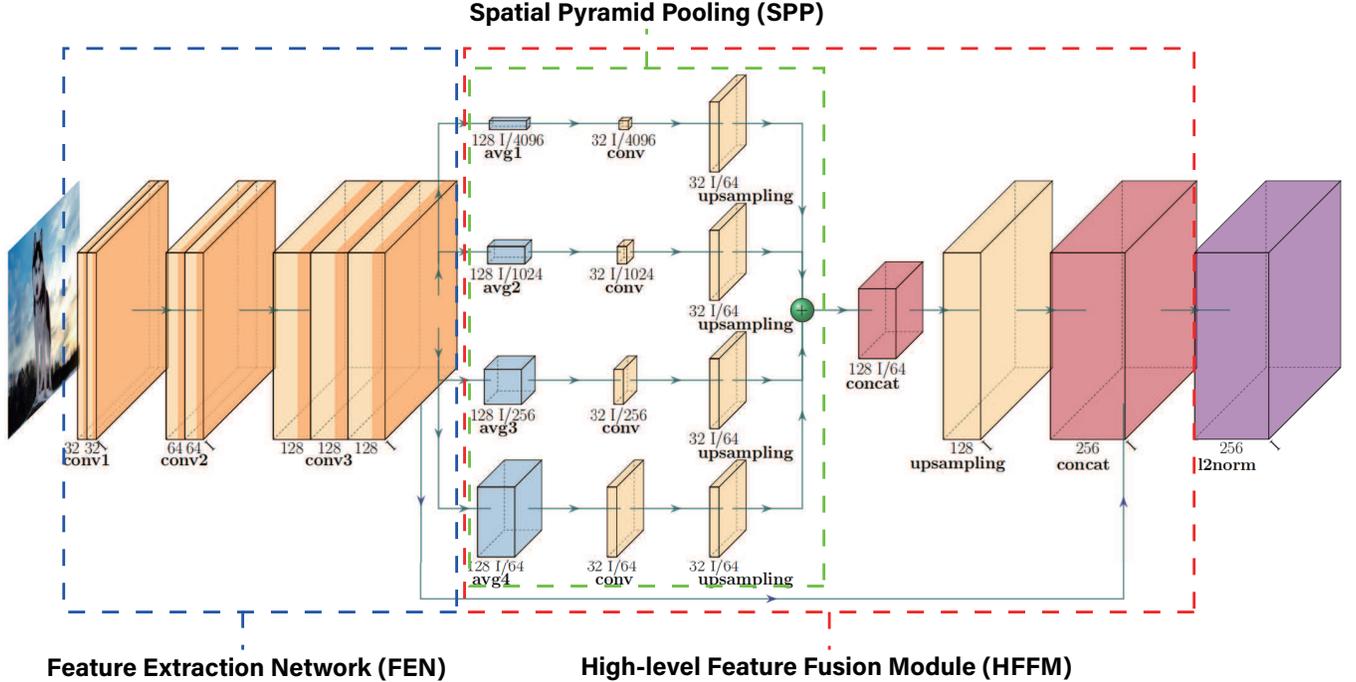}}
\end{minipage}
\caption{The architecture of the proposed network. An image is fed to the pipeline consisting of a feature extraction network (indicated by the blue dotted box) for low-level information extraction around the point, and a high-level feature fusion module (indicated by the red dotted box) for integrating the high-level information and the low-level information. In order to extract high-level information, we use a spatial pyramid pooling (indicated by the green dotted box) module to obtain the information at different scales around the point. Finally, the output of the descriptors is obtained through an L2 normalization layer.}
\label{fig:res}
\end{figure*}

Many learning-based approaches hope to obtain the coordinates or descriptors or both of them of the keypoints with a data-driven approach. MatchNet~\cite{14} is an early example. It uses a similarity measure network to calculate the similarity of descriptors, which significantly improves previous results. Later LIFT~\cite{15} is the first completely learning-based end-to-end network that outputs both keypoints coordinates and descriptors. DELF~\cite{16} is applied to the image retrieval task of street view markers, roughly locating the keypoints on the image through an attention mechanism. SuperPoint~\cite{17} is based on the synthetic datasets, which also outputs keypoints coordinates and descriptors at the same time. D2-Net~\cite{18} obtains keypoints coordinates and descriptors by selecting soft local-max in a pre-trained network. These learning-based methods, which obtain the coordinates and descriptors at the same time, generally adopt multi-task optimization to train the network, which imperceptibly increases the training difficulty of the network. All of the above methods need some custom extremum to obtain the feature points of the texture region. Moreover, most of the methods focus on the local features of keypoints and do not consider the high-level information around the keypoints.

According to common sense, there is no key point in flat regions, like points C and D shown in Fig.1. For the points C and D with a small neighbor (red circles) neither give valid descriptions. With the increase of the receptive field, it is possible to find valid descriptions for points C and D with a larger neighbor (green circles). Therefore, we propose a descriptor extraction network that fuses more information which contains the features of the texture area in the green circle (fusion of the high-level information and low-level information around the keypoints), makes points C and D divisible, and ensures that points in the flat region can also be well matched. 

Our contributions can be summarized from two aspects:

1. We introduce a simple uniform sampling method for keypoint extraction to enlarge the candidate range of keypoints, which can obtain a larger distribution of the matching feature point pairs accompanying by more abundant image information.

2. We propose a region-based descriptor network (RDN) that integrates high-level information and low-level information around the points, which has better expressive power and even makes the points in the flat region can be distinguished from each other.

\section{PROPOSED METHOD}
\label{sec:pagestyle}

\subsection{Motivation and Our Ideas}
\label{ssec:subhead}

In general, the coordinates and the descriptors of the keypoints need to be obtained firstly for further matching. Most of the previous methods used various extremum points as feature points, which are concentrated in the region with rich texture. In this way, the points out of the rich texture regions are usually not considered as candidates of a matching point. We hope to integrate more comprehensive image information so that flat areas can also have feature points to participate in matching.

\subsubsection{Keypoints Extraction}
\label{sssec:subsubhead}

In many tasks, the keypoint does not have a unified definition, and whether it is an extreme point does not affect the result of matching, such as in the calculation of the fundamental matrix~\cite{37}. As long as a sufficient number of point pairs match correctly, we can calculate the relationship between the images, such as the rotation and translation of the image perspective. Therefore, we came up with a uniform sampling method on the image to extract keypoints. 

\subsubsection{Learning Descriptors}
\label{sssec:subsubhead}

As shown in Fig.1, the local features extracted by the previous methods cannot well accomplish the matching of feature point pairs in flat regions, that is, the descriptors of feature points in flat regions are indistinguishable. So we propose a fusion of high-level information and low-level information convolutional neural network to address this problem.

\subsection{Network Architecture}
\label{ssec:subhead}

The proposed network, referred to as RDN, aims to generate valid descriptors for points of an input image even in flat regions. As shown in Fig.2, the input is an image $I\in R^{H\times W\times 3}$. The front part of the RDN is the feature extraction network (FEN) which is inspired by L2-Net~\cite{19} and R2D2~\cite{36}, and the latter part is a high-level feature fusion module (HFFM). The 128-dimensional vector generated by the FEN represents the low-level features ($d_{low}\in R^{H\times W\times 128}$) around the keypoints. Then it serves as the input to the HFFM. The output of the HFFM is the high-level features ($d_{high}\in R^{H\times W\times 128}$) around the keypoints. To effectively captures contextual relations around the keypoints, the HFFM starts with four fixed-size average pooling blocks for Spatial Pyramid Pooling (SPP): $64\times 64, 32\times 32, 16\times 16$, and $8\times 8$. Then the feature maps are upsampled to the same size as the input figure. By concatenating the outputs of the FEN ($d_{low}$) and the HFFM ($d_{high}$), we can obtain a 256-dimensional vector, for each pixel. To better compare the descriptors, an L2 normalization layer was added to normalize the descriptors. The final output are the dense descriptors $d_{output}\in R^{H\times W\times 256}$.

\subsection{Training Loss}
\label{ssec:subhead}

In terms of descriptor learning, the descriptors are expected to be distinctive and be avoiding mismatching. We adopt the triplet loss similar to D2-Net, which has been successfully applied in descriptors learning \cite{26}. The RDN does not need to predict the coordinates of keypoints, so we remove the weighted part of loss calculation in D2-Net, and directly calculate the triplet loss between the distance of the descriptors of the matching keypoints and the descriptors of the 
\begin{figure}[htb]
\begin{minipage}[b]{1.0\linewidth}
  \centering
  \centerline{\includegraphics[width=8.5cm]{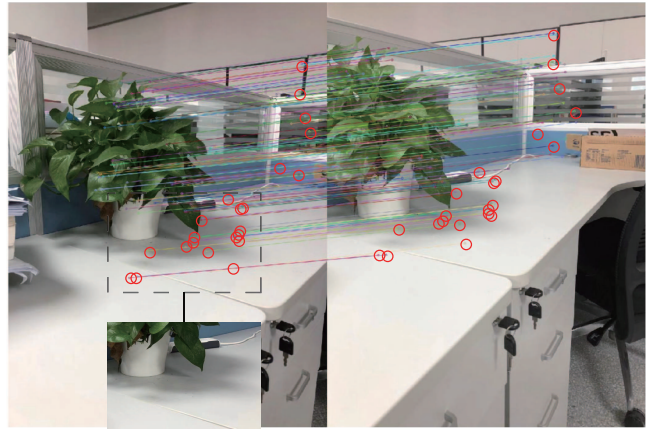}}
\end{minipage}

\caption{Feature points on flat areas of the table  can also be matched as shown in the black dotted box.}
\label{fig:res}
\end{figure} 
mismatched keypoints. It can be formulated as follows:
\begin{equation}
Loss(I_1,I_2)=\sum_{c\in C}f(p(c),n(c)),
\end{equation}
where $I_1$, $I_2$ are two images with the same scene, f indicates the triplet margin ranking loss, c denotes the correspondences, p represents the distance between the descriptors of the matched keypoints pair (correspondence), n is the nearest distance between the descriptors of the mismatched keypoints pair, C is the set of the correspondences in the two images. See \cite{18} for more details.

\begin{figure*}[htb]
\begin{minipage}[b]{1.0\linewidth}
  \centering
  \centerline{\includegraphics[width=18.5cm]{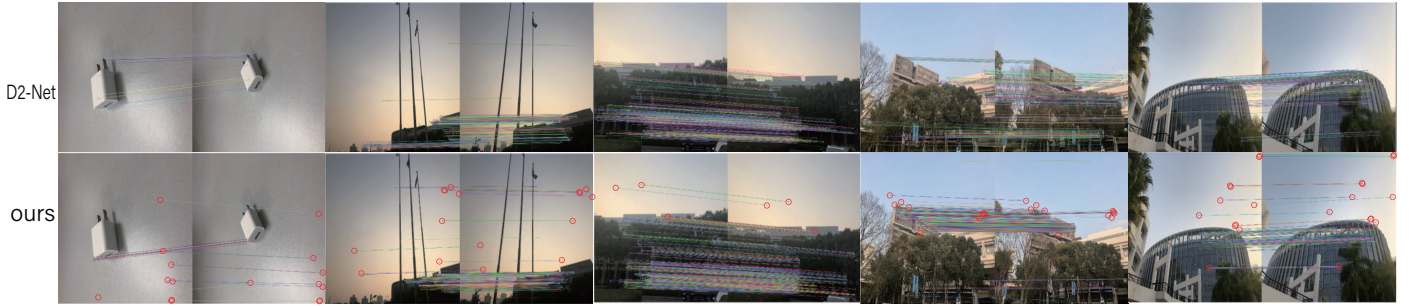}}
\end{minipage}
\caption{Matching results of different scenarios.}
\label{fig:res}
\end{figure*}

\section{EXPERIMENTS}
\label{sec:typestyle}

\subsection{Dataset and Implementation Details}
\label{ssec:subhead}

To generate training data close to the real scenario, we use the MegaDepth Dataset \cite{20} from the internet photo gallery. Since D2-Net has extracted the pixel-wise correspondences with the COLMAP \cite{1,22}, we directly use it as ground truth to train our network.

The architecture similar to L2-Net, pretrained on the Oxford and Pairs retrieval dataset \cite{27} and the Aachen Day-Night dataset \cite{28}, was used to initializes the FEN. The HFFM was fine-tuned for 50 epochs using Adam with an initial learning rate $10^{-3}$, which was further divided by 2 every 10 epochs. 100 images were randomly selected from each scene for training to solve the problem of the unbalanced number of images in the scene. The size of the training batch is 1. 

\subsection{Qualitative and Visual Evaluation}
\label{ssec:subhead}

\noindent To verify the effectiveness of our proposed method, we conduct some experiments on pictures taken with a mobile phone. As we can see in Fig.3, not only textured areas have many matching points, but flat areas, such as tabletop, cabinet surface, and glass surface, can also get matching keypoints.

Our method is completed by referring to D2-Net, so we carry out some experiments comparing with D2-Net. D2-Net and our method are used to get the feature points and descriptors respectively in the experiment. The NNS algorithm is used to match the feature points, and then RANSAC is used to screen out the outliers. The final result is shown in Fig.4. We can see that D2-Net matches basically the points in the texture region, and our method has significantly more matching point pairs in the flat region than D2-Net. Our method can preserve more information about the image and introduce less error for subsequent advanced tasks.

\subsection{Quantification and Camera Localization}
\label{ssec:subhead}

\begin{table}[!htbp]
\centering
\caption{Results on the Aachen Day-Night dataset, we report the percentages of images that are successfully located under the 3 error thresholds. The best performance is shown in bold.}

\begin{tabular}{lccc}
\hline
Method& 0.5m,$2^{\circ}$& 1m,$5^{\circ}$& 5m,$10^{\circ}$\\
\hline
RootSIFT& 33.7& 52.0&65.3\\
HAN+HN& 37.8& 54.1& 75.5\\
SuperPoint& 42.8& 57.1&75.5\\
DELF& 39.8& 61.2&85.7\\
D2-Net& 44.9& 66.3& {\bf 88.8}\\
R2D2& 45.9& 67.3&{\bf 88.8}\\
\hline
{\bf RDN(ours)}& {\bf 65.3}& {\bf 75.5}& 84.7\\
\hline
\end{tabular}
\end{table}

\noindent We evaluate our approach in a camera localization task. The dataset that we use is the Aachen Day-Night dataset with dramatic changes in illumination. This dataset has night-time images with unknown camera pose and day-time images with a known camera pose. The day-time images need to be reconstructed first to obtain the 3D scene structures, and then the night-time query images should be located from these 3D structures. We follow the evaluation protocol \cite{28} and report the percentage of successful camera localization obtained from night-time image queries within three error threshold (0.5m,$2^{\circ}$)/(1m,$5^{\circ}$)/(5m,$10^{\circ}$). The camera pose error consists of the camera position (meter) and the camera orientation (degree). We compare the performance of the proposed approach with other methods including RootSIFT, HardNet++ (HAN) with HesAffNet features (HN), SuperPoint, DELF, D2-Net, R2D2. In addition to our method, the data in Table 1 directly adopt the data of \cite{36}.

As we can see in Table 1, our approach achieves the best result under strict threshold conditions for camera pose estimation. By means of uniform sampling, our method retains more regional information, and meanwhile, by integrating more context information, obtains a more comprehensive description of point features and better express the information contained in the image. Therefore, our method can achieve the most accurate camera positioning. The reason why the effect is not good under the condition of a large error threshold is that the results obtained by our method in the reconstructed the 3D model and feature extraction of the night-time images mainly focus on the high-precision part, while the parts with low-precision are relatively few, so the influence of relaxing threshold is not as good as  other methods. The results show that our simple approach can be used for estimating accurate camera pose.

\section{CONCLUSIONS}
\label{sec:majhead}

In this paper, we introduce a new method to obtain keypoints without extremum operation, which significantly increases the range of keypoints acquisition. This may lead to a problem that points in a flat region have no locally meaningful features. To address this problem, we propose a region-based descriptor by combining the context features of a deep network. We obtain more expressive descriptors by combining high-level information and low-level information. Compared to our baseline D2-Net, we obviously obtain keypoints pairs of more regions that could express more information about the images. Moreover, our approach achieves the best performance in terms of camera positioning by collecting more information from the images under a strict threshold. 



\bibliographystyle{IEEEbib}
\bibliography{refs}

\end{document}